\PassOptionsToPackage{table}{xcolor}
\documentclass[]{academic_template}
\usepackage[toc,page,header]{appendix}

\usepackage{marvosym}

\usepackage{bbding}
\usepackage{soul}

\usepackage{amssymb}
\usepackage{amsmath}

\usepackage{makecell}
\usepackage{algorithm}      
\usepackage{algpseudocode}  
\usepackage{tabularx}
\usepackage{booktabs}
\usepackage{graphicx}
\usepackage{array}
\newcolumntype{C}{w{c}{3em}}
\usepackage{listings}
\usepackage{hyperref}
\usepackage{url}
\usepackage{booktabs}
\usepackage{wrapfig}

\usepackage{xcolor}

\usepackage{colortbl}
\usepackage{caption}
\usepackage{subcaption}

\usepackage{amsthm}

\setlogoheight{8mm}   

\setlogospacing{0mm}
\setsjtublue 

\settitlerulethickness{3pt}  

\abstractboxon 
\setabstractframecolor{gray} 
\setabstractbgcolor{gray!10} 
\setlogotolineshift{0mm} 

\settoprulethickness{2.5pt}  
\setbottomrulethickness{1.5pt} 

\usepackage{dblfloatfix}
\usepackage{float}

\usepackage{tabularx}

\usepackage{multicol}
\usepackage{multirow}
\usepackage[table]{xcolor}  

\usepackage{natbib}
\setcitestyle{numbers}
\setcitestyle{square}
\usepackage{graphicx}

\usepackage{xspace}

\title{ACE-Brain-0.6: Scaling Spatial Grounding for \\
Physical Foundation Models}

\author{%
\begin{minipage}{\textwidth}
    \centering
    \sffamily
    \setlength{\parskip}{0pt}

    {\fontsize{12}{16}\selectfont
    ACE-Brain Team\par}

    \vspace{1.2em}

    {\fontsize{10}{14}\selectfont
    Please see
    \hyperref[app:contributors]{Contributions and Author List}
    for more author details.\par}
\end{minipage}%
}

\abstract{
General-purpose multimodal models, including GPT-6 Astra and Claude Opus 5.5, demonstrate broad capabilities in embodied manipulation.
Our analysis indicates that this generality is accompanied by strong spatial perception, as Astra can determine where to interact with task-relevant entities amid visual distractors.
Because spatial perception connects high-level reasoning to the physical scene, we investigate scaling visual grounding to strengthen the perceptual foundation of general embodied operation.
We develop a unified data pipeline that combines robot manipulation recordings with complementary visual data from diverse sources.
After automatically annotating the robot recordings, the pipeline converts the collected data into verifiable grounding supervision.
Fine-tuning Qwen3.5-9B on approximately 1.5 million samples produced by this pipeline yields ACE-Brain-0.6.
The model outperforms GPT-6 Astra on four of seven public grounding benchmarks, with an overall average score of 75.41 compared with Astra's 73.68.
Under a fixed Qwen3.5-9B fine-tuning setup, aggregate grounding performance improves steadily as successive data releases expand the training collection to cover a broader range of grounding scenarios.
These results show that scaling verifiable grounding supervision from diverse sources can strengthen the spatial perception of an open 9B model for general embodied operation.
}
\date{\today}

\newif\ifusebibliography
\usebibliographyfalse

\begin{document}

\phantomsection
\addcontentsline{toc}{section}{Abstract}
\maketitle

\newpage
\tableofcontents

\newpage
\section{Introduction}
\label{sec:introduction}
General-purpose reasoning models are beginning to generalize across diverse embodied settings.
GPT-6 Astra~\cite{openai2026gpt6astra}, in particular, demonstrates strong performance in embodied manipulation.
A closer examination of its behavior shows that this embodied competence is accompanied by strong spatial perception.
It can localize task-relevant interaction targets in cluttered scenes despite the presence of visually similar distractors.

This observation highlights the role of spatial perception in translating high-level reasoning into physical action.
Grounding provides this connection by linking natural-language descriptions to their visual referents in an image~\cite{kamath2021mdetr,shridhar2021cliport}.
Reliable grounding requires precise localization of all image regions that satisfy a description, with explicit recognition when no match exists~\cite{liu2023gres}.
In embodied environments, clutter and occlusion make this correspondence particularly difficult to establish~\cite{wang2021ocid}.
A grounding error can therefore misdirect an agent's actions even when its high-level reasoning is sound.

\begin{figure}[!t]
    \centering
    \includegraphics[width=\linewidth]{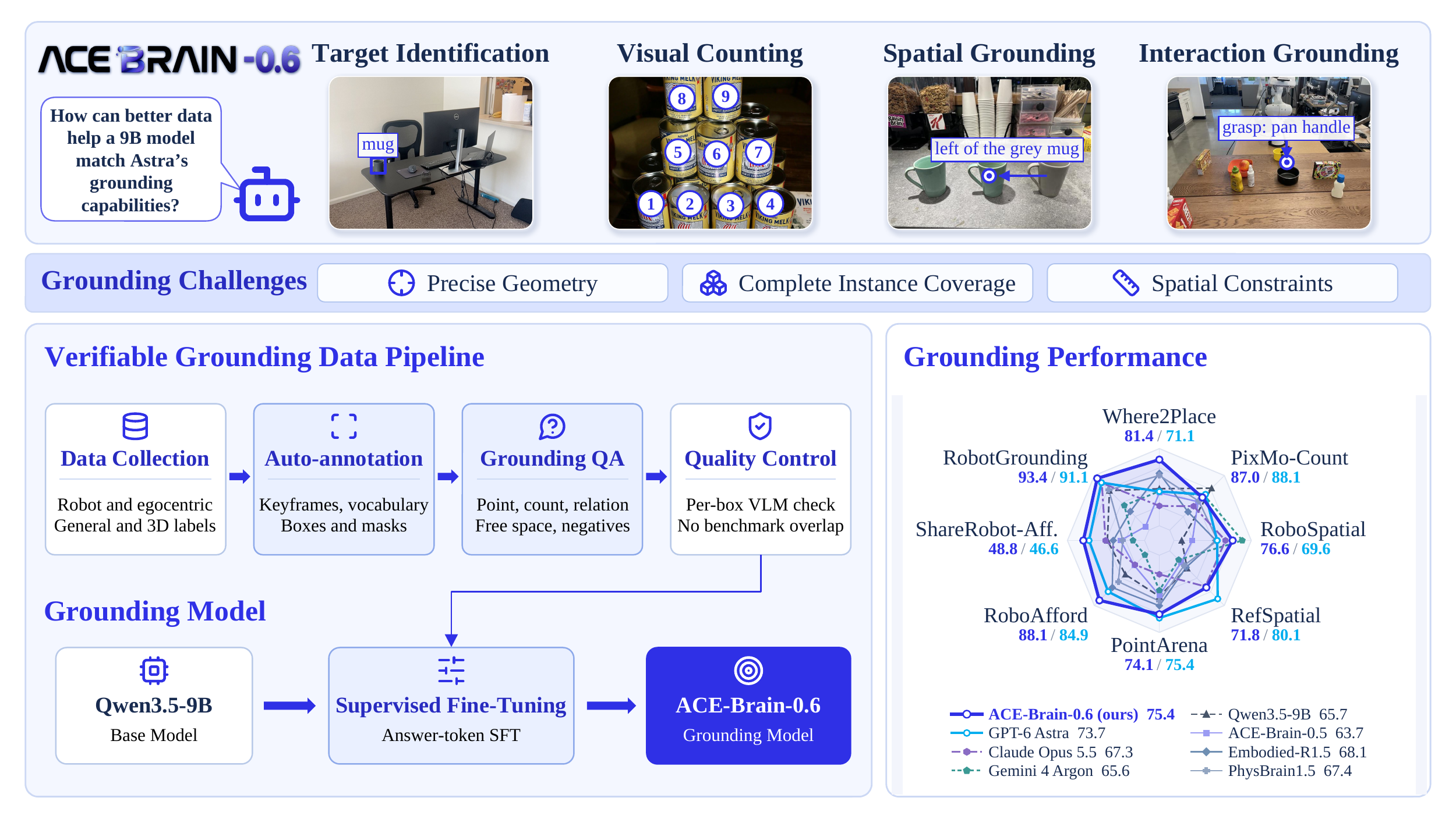}
    \caption{
        \textbf{Overview of ACE-Brain-0.6.}
        Our data pipeline converts diverse visual sources into
        verifiable grounding supervision for fine-tuning Qwen3.5-9B.
        The resulting model supports target identification, visual
        counting, spatial grounding, and interaction grounding.
        The radar chart compares performance across eight benchmarks.
    }
    \label{fig:overview}
\end{figure}

We therefore examine Astra's spatial grounding capabilities in greater detail.
Our analysis in Section~\ref{sec:astra-grounding-analysis} illustrates its ability to handle diverse grounding queries, while also revealing localization errors in tasks with precise spatial requirements.
Using Astra as a reference, we ask whether an open 9B-parameter vision-language model can match or surpass Astra's grounding performance by expanding the scale and coverage of grounding supervision, with the model architecture and training recipe held fixed.

To investigate this question, we develop a unified pipeline for constructing grounding supervision from existing open datasets.
The pipeline combines robot manipulation recordings from more than thirty dataset families with complementary visual data to broaden task coverage.
We automatically annotate the robot recordings to supply the object-level labels they generally lack.
All sources are then converted into question--answer supervision that is verifiable against the supporting annotations, with controls to reduce shortcuts between question wording and answer format.
This pipeline yields approximately 1.5 million training samples, which we use to fine-tune Qwen3.5-9B~\cite{qwen3.5} for a single epoch, producing ACE-Brain-0.6.
Figure~\ref{fig:overview} provides an overview of our approach.

We evaluate ACE-Brain-0.6 on seven public grounding benchmarks.
The model outperforms the Qwen3.5-9B baseline on six benchmarks and remains behind on PixMo-Count.
It exceeds GPT-6 Astra on four benchmarks and achieves a slightly higher seven-benchmark average (75.41 versus 73.68), while remaining behind on PixMo-Count, RefSpatial, and PointArena.
These results suggest that a 9B open model trained with diverse, verifiable grounding supervision can match Astra's aggregate performance on the evaluated benchmarks.

We further examine data scaling by fine-tuning from the same base checkpoint with a fixed training recipe for each data release.
Across releases that expand both sample count and supervision coverage, the seven-benchmark average increases approximately linearly with the logarithm of sample count over the evaluated range, slightly exceeding Astra's average at about 1.5 million samples.
In a control experiment, adding samples only from existing sources improves this average by less than one point, suggesting that broader supervision coverage contributes to the observed gains.

Our contributions are threefold:
\begin{itemize}
    \item \textbf{A data pipeline for grounding.} We collect large-scale open-source robot, egocentric, object-centric, and spatial data, automatically annotate robot recordings, and convert these sources into verifiable question--answer supervision for target identification, complete instance coverage, and precise localization.
    
    \item \textbf{An empirical study of grounding-data scaling.} Through a controlled series of training runs, we show that grounding performance improves steadily with data scale and that expanding the coverage of supervision contributes substantially more than increasing the volume of existing sources.
    
    \item \textbf{ACE-Brain-0.6.} By scaling verifiable grounding data to approximately 1.5 million samples, we train a 9B model that exceeds GPT-6 Astra on four of seven grounding benchmarks and achieves a higher overall average.
\end{itemize}

\section{Related Work}
\label{sec:related_work}

\subsection{Visual Grounding and Embodied Spatial Understanding}

Visual grounding links language descriptions to image locations. Detector-based methods such as GLIP and Grounding DINO align visual and language features for phrase grounding and open-set localization~\cite{li2022,liu2024}. Generative approaches represent spatial outputs as token sequences, from coordinate and class prediction in Pix2Seq~\cite{chen2022pix2seq} to multi-task visual prediction in Florence-2~\cite{xiao2023} and detection, referring, and pointing in Rex-Omni~\cite{jiang2025}. Grounding also supports broader embodied understanding: PaLM-E integrates sensor inputs into language modeling~\cite{driess2023palme}, Gemini Robotics-ER develops spatial understanding and embodied reasoning~\cite{geminirobotics2025}, and RynnBrain and Embodied-R1.5 combine localization with embodied understanding and reasoning~\cite{dang2026,yuan2026}. These works connect spatial prediction with increasingly broad model capabilities. We focus on grounding itself, keeping the base vision-language model and training recipe fixed to examine the effect of training data.

\subsection{Grounding Data and Supervision}

Grounding supervision comes from human annotation, automatic labeling, and geometric derivation. Molmo-and-PixMo collects language-conditioned point annotations, including complete instance sets and absent-target examples~\cite{deitke2024molmopixmoopenweights}. GrIT pairs caption phrases with predicted image regions~\cite{peng2023}. SpatialVLM constructs spatial question--answer data using segmentation, depth, and object descriptions~\cite{chen2024}, while RoboPoint derives instruction-point supervision from simulated scene geometry~\cite{yuan2024robopoint}. We build on these approaches by combining robot recordings, egocentric interactions, dense object and part annotations, and 3D scenes. Our pipeline automatically annotates robot recordings and converts these heterogeneous sources into a common question--answer format. Answers remain linked to supporting annotations, with checks for instance completeness, spatial constraints, and target absence. This training collection allows us to study how supervision scale and composition affect grounding performance.

\section{Analysis of GPT-6 Astra}
\label{sec:astra-grounding-analysis}

Our experiments with GPT-6 Astra (hereafter Astra) as a robotic manipulation policy highlight how visual grounding supports analytic geometry in translating task objectives into physical actions. Grounding identifies relevant objects, interaction regions, and spatial references in the observed scene, and analytic geometry uses spatial quantities derived from these references to determine positions, relative displacements, and motion parameters. In this section, we first illustrate this dependency through two manipulation cases, and then examine Astra's grounding capabilities in greater detail.

\subsection{Case Studies: From Grounding to Geometric Robot Manipulation}
\label{sec:astra-manipulation-cases}

\begin{figure*}[!htb]
    \centering
    \includegraphics[width=\textwidth]{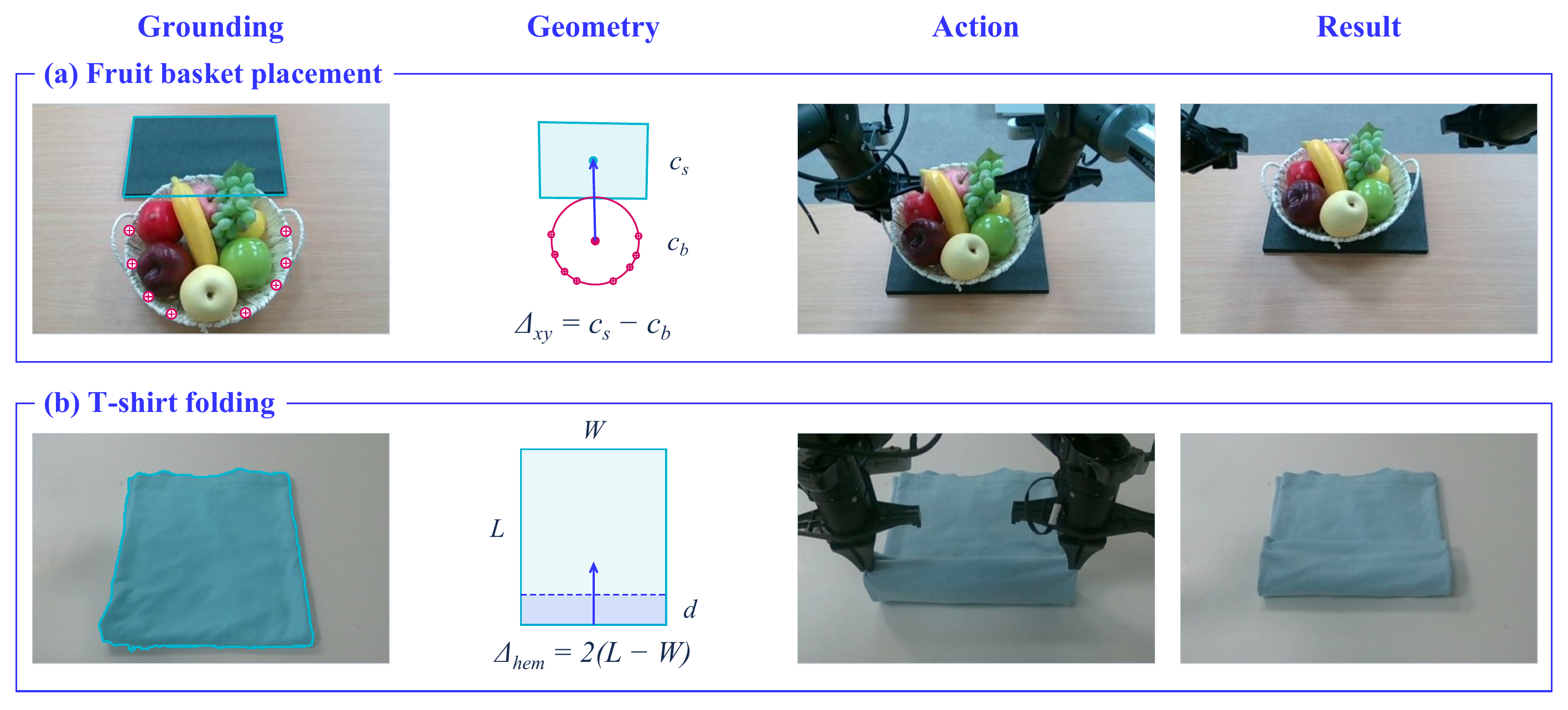}
    \caption{\textbf{Grounding supports geometric manipulation.}
    (a) Basket-rim points and the destination region provide spatial references for placement.
    (b) A garment mask and contour provide extent estimates for selecting a square fold.
    Magenta marks selected points; cyan marks regions or contours.
    Overlays and schematics reconstruct recorded calculations; cropped RGB frames show execution and outcomes after release.
    The T-shirt mask was obtained through Astra-authored color segmentation.}
    \label{fig:astra_manipulation_cases}
\end{figure*}

The two cases illustrate complementary forms of grounding used in geometric manipulation (Figure~\ref{fig:astra_manipulation_cases}). Fruit basket placement requires selecting points on the manipulated object and its destination, whereas T-shirt folding requires extracting the garment's spatial extent. In both cases, these visual references determine the geometric quantities used to construct the action.

\par\medskip
\noindent\textbf{Fruit basket placement.}\quad
To place a loaded fruit basket onto a black sponge, Astra selected points along the basket rim and the corners of the destination surface. Using calibrated RGB-D observations, it fitted the basket rim and support plane to estimate their relative configuration. The resulting planar centers of the basket and destination, $\mathbf{c}_b$ and $\mathbf{c}_s$, defined an initial displacement $\boldsymbol{\Delta}_{xy}=\mathbf{c}_s-\mathbf{c}_b$. Astra used these estimates to construct coordinated arm motions and reassessed the basket's pose during placement. Here, grounding specifies the object geometry and support surface to which the placement calculation applies.

\par\medskip
\noindent\textbf{T-shirt folding.}\quad
To fold an already side-tucked T-shirt into a square, Astra constructed a color-based mask within a selected image region and extracted the garment contour. Projecting this contour onto an assumed cloth plane gave an estimated length $L\approx363\,\mathrm{mm}$ and width $W\approx300\,\mathrm{mm}$. These dimensions informed a bottom fold of depth $d=L-W$, with a corresponding hem displacement $\Delta_{\mathrm{hem}}=2d\approx125\,\mathrm{mm}$. Both arms executed the staged fold, and Astra extracted the contour again after release to assess the resulting shape. Grounding the garment's extent thus supplies the dimensions needed to select a fold that satisfies the task's shape requirement.

Together, these cases illustrate how task-relevant grounding supplies the spatial inputs needed for geometric action construction and subsequent visual verification. These observations motivate studying grounding as a measurable component of spatial perception and using Astra's capabilities as a reference for improving our model.

\subsection{A Closer Look at Astra's Grounding Capabilities}
\label{sec:astra-grounding-capabilities}

\begin{figure*}[!htb]
    \centering
    \includegraphics[width=\textwidth]{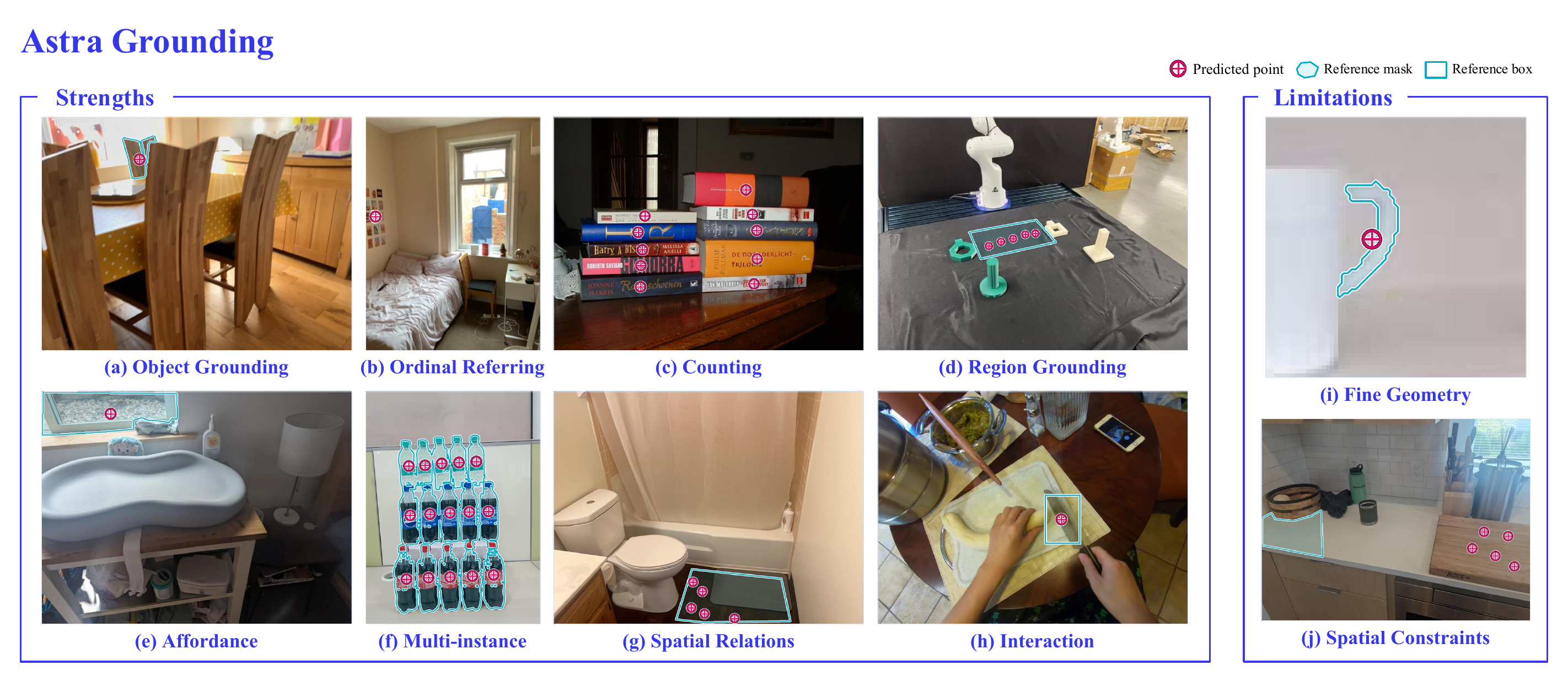}
    \caption{\textbf{Strengths and limitations of Astra's grounding.}
    (a--h) Examples of object and ordinal referring, counting and multi-instance localization,
    spatial and region grounding, and functional and interaction grounding.
    (i) A prediction near the mug handle falls outside the reference mask.
    (j) Predicted points fall outside the reference region in front of the steamer.
    Magenta markers indicate Astra's predicted points; cyan masks and boxes denote reference annotations.}
    \label{fig:astra_grounding}
\end{figure*}

Figure~\ref{fig:astra_grounding} illustrates Astra's grounding through eight examples of its strengths and two examples of its limitations. The queries require selecting objects, locating all members of a set, identifying spatial regions, or associating functions and actions with visual locations. These examples make concrete the range of visual references that Astra can supply for subsequent geometric reasoning and the localization errors that can arise.

\par\medskip
\noindent\textbf{Strengths.}\quad
Astra can resolve descriptions that distinguish a particular instance from other objects of the same category. In the object grounding example (a), it identifies the chair farthest from the viewer. In the ordinal referring example (b), it selects the picture that is third from the top in the second column from the left.

Grounding also extends to sets of instances. In the counting example (c), Astra returns ten points corresponding to ten books, with the number of predictions matching the reference count. In the multi-instance example (f), it returns a point on each of the 15 bottles. Both queries ask for all matching objects, and both responses represent the target set through individual locations.

The examples of region grounding and spatial relations require locating unoccupied space under relational descriptions. In (d), Astra places points in the free space between the green and white blocks at the back of the scene. In (g), it locates vacant space to the right of the toilet. The predictions lie within the corresponding reference regions.

Functional descriptions and interaction goals provide further ways to specify a visual target. In the affordance example (e), Astra locates the window from a description of a structure that can be opened or closed to admit light and air. For the interaction task in (h), it predicts a point within the annotated end-effector interaction region for cutting with a knife. The former identifies an object through its function; the latter localizes a region relevant to a specified action.

\par\medskip
\noindent\textbf{Limitations.}\quad
The fine-grained localization example (i) asks which part of a mug should be gripped to lift it. Astra's point lies close to the handle but just outside the valid reference mask. The prediction is near the relevant object part, yet it fails the geometric localization requirement. This case highlights the sensitivity of fine-grained grounding to local boundaries.

The spatially constrained query in (j) asks for free space in front of the steamer, but Astra places its points on the cutting board to the right, outside the annotated region. Here, the error concerns localization of the region specified by a spatial relation. Together, the two failures show why grounding must satisfy both local geometric boundaries and relations to the intended reference objects.

\par\medskip
\noindent\textbf{Implications for ACE-Brain.}\quad
These examples illustrate a range of grounded references that can support geometric reasoning, while the failures expose remaining difficulties in precise localization and spatial constraint satisfaction. They motivate using Astra's grounding performance as a reference for developing these capabilities in our model. We investigate whether scaling diverse, verifiable grounding supervision can bring our model toward or beyond this reference. Section~\ref{sec:method} describes the data pipeline, and Section~\ref{sec:exp} evaluates the resulting gains and remaining gaps across grounding tasks.

\section{Method}
\label{sec:method}

This section describes the three components of ACE-Brain-0.6: the model and the grounding tasks it is trained to solve (Section~\ref{sec:method:model}), the data we collect and the pipeline that turns it into grounding supervision (Section~\ref{sec:method:data}), and the training recipe (Section~\ref{sec:method:training}).

\subsection{Model and Problem Formulation}
\label{sec:method:model}

\par\medskip
\noindent\textbf{Grounding as text generation.}\quad
 We cast every grounding task as conditional text generation: given one or more images and a natural-language query, the model produces a short answer in one of five forms, namely a point $(x, y)$, a list of points, a box $(x_1, y_1, x_2, y_2)$, an integer count, or a categorical token (\emph{yes} or \emph{no} for spatial judgments, \emph{none} for absent targets). Coordinates are integers normalized to $[0, 1000]$ along each image axis, and point lists are written as $[[x_1, y_1], \ldots]$ without per-point labels. This single output interface covers the task families behind the capabilities of Section~\ref{sec:astra-grounding-analysis}: referring by category, attribute, position, relation, or function; pointing to and counting every instance of a set; spatial relation judgments and free-space localization; grounding of parts, contact regions, placement destinations, and end-effector trajectories; and, in robot scenes, locating a target across two camera views.

\par\medskip
\noindent\textbf{Base model.}\quad
ACE-Brain-0.6 is initialized from Qwen3.5-9B~\cite{qwen3.5}. We fine-tune all language-model parameters, keep the vision encoder frozen, and disable the multi-token-prediction head. The model is trained and evaluated in non-thinking mode: the chat template emits an empty reasoning block before each answer, and we exclude these template tokens from the training loss so that the model never learns to emit them inside an answer.

\subsection{Data Pipeline and Data Scale}
\label{sec:method:data}

\par\medskip
\noindent\textbf{Data collection.}\quad
We draw training data from three kinds of sources. \emph{Robot manipulation data} covers the target domain: manipulation episodes spanning single-arm, dual-arm, mobile, and humanoid embodiments. Since these recordings rarely carry object labels, we annotate them automatically as described below. \emph{Egocentric human--object interaction data} adds first-person views close to a wrist or head camera, with hand--object masks and contact regions. \emph{General pointing and grounding data} supplies what robot data lacks: exhaustive instance sets for counting, part masks for functional grounding, and 3D annotations for spatial relations and free space; a small amount of general visual question answering data preserves broad multimodal ability.

\begin{figure}[!htb]
    \centering
    \includegraphics[width=\linewidth]{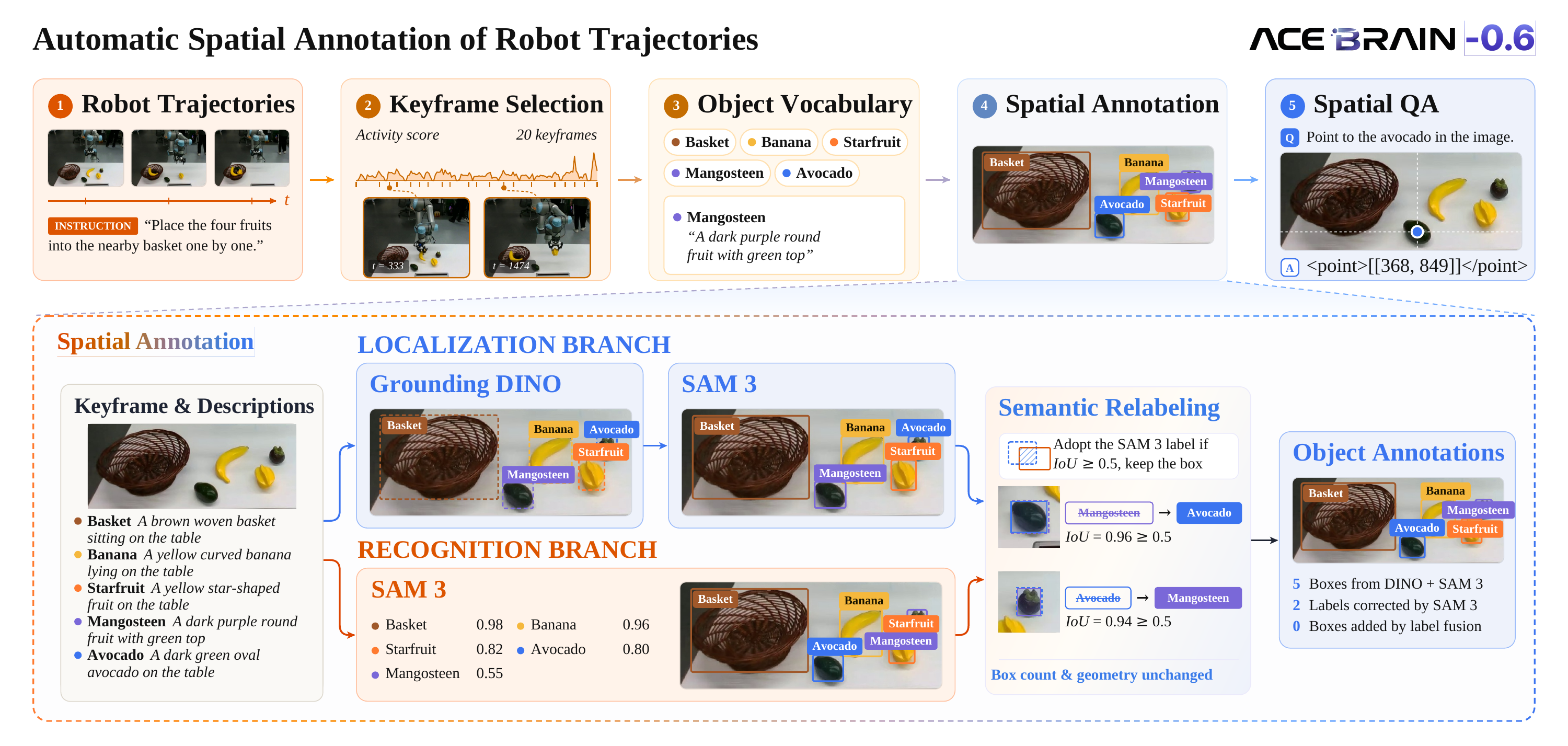}
    \caption{
        \textbf{Automatic spatial annotation of robot trajectories.}
        The pipeline selects keyframes, constructs an object
        vocabulary for each episode and camera, and produces spatial annotations
        for generating question--answer pairs.
        Localization and recognition branches are combined through
        semantic relabeling, which corrects object labels while
        preserving the number and geometry of the bounding boxes.
    }
    \label{fig:annotation_pipeline}
\end{figure}

\par\medskip
\noindent\textbf{Automatic annotation of robot data.}\quad
Robot recordings provide observations, actions, and instructions but rarely object labels. We convert them into grounding supervision in three steps (Figure~\ref{fig:annotation_pipeline}). First, we select a few keyframes per episode from the action signal, favoring gripper-state transitions and large action changes while keeping the start, middle, and end of the trajectory. Second, a VLM lists the objects visible in each episode and camera with short names and appearance descriptions~\cite{patzold2025leveraging}, giving an open vocabulary tied to the scene and viewpoint. Third, Grounding DINO~\cite{liu2024} detects these objects and SAM~3~\cite{carion2026sam} refines their boxes; a separate SAM~3 recognition branch then relabels each box with its best-overlapping concept detection, correcting names without moving or adding boxes. The resulting annotations yield robot-scene supervision for referring, pointing, counting, free-space and destination grounding, and rejection of absent objects.

\par\medskip
\noindent\textbf{Question generation.}\quad
All sources are turned into question--answer pairs from templates, under rules that keep every answer verifiable against its annotation. Counting and every-instance questions are emitted only when the instance set is complete; negative questions are checked against the vocabulary of the image so that none is correct; free-space targets mark unoccupied 2D regions and do not assert 3D feasibility. Answers are written in the output interface of Section~\ref{sec:method:model}, and shortcuts between question wording and answer form are removed: no answer form exceeds 70\% within a template, yes/no answers are balanced, and the number of points matches what the question asks for.

\par\medskip
\noindent\textbf{Quality control and data scale.}\quad
Every detector box in robot scenes is verified by a VLM and dropped if its label does not match the region. Images and episodes from the evaluation benchmarks are excluded from training by matching. One training sample is one or more images with a conversation of one or more question--answer turns. The final collection contains about 1.54 million samples, comprising about 3.41 million question–answer pairs across about 1.39 million distinct image paths.

\subsection{Training}
\label{sec:method:training}

\par\medskip
\noindent\textbf{Objective.}\quad
We minimize the token-averaged cross-entropy over answer tokens only. Question and image tokens receive no loss, and, as noted in Section~\ref{sec:method:model}, the four tokens of the empty reasoning block are masked as well. Conversations are packed into fixed-length sequences of 16{,}384 tokens with variable-length attention, so that tokens from different conversations never attend to each other; each global step consists of 16 such sequences, or roughly 140 samples.

\par\medskip
\noindent\textbf{Optimization.}\quad
We use AdamW with $\beta = (0.9, 0.95)$, $\epsilon = 10^{-8}$, and weight decay $0.1$. The learning rate warms up linearly over 50 steps to a peak of $1 \times 10^{-5}$ and then follows a cosine schedule down to $1 \times 10^{-7}$. The peak learning rate was selected once by a four-point sweep on an early data release and then held fixed for every subsequent run, so that differences between runs reflect differences in data alone. Each run trains for a single epoch over its data. Master weights are kept in fp32, and training uses fully sharded data parallelism with activation checkpointing and FlashAttention-2.

\section{Experiments}
\label{sec:exp}

We compare ACE-Brain-0.6 with open embodied models of the same size, and with three closed models, GPT-6 Astra, Claude Opus~5.5 and Gemini 4 Argon on seven public grounding benchmarks and on a robot-scene benchmark of our own (Sections~\ref{sec:exp:setup} and~\ref{sec:exp:main}). We then ask where the gains come from (Section~\ref{sec:exp:capability}), and how they grow with the training data (Section~\ref{sec:exp:scaling}).

\subsection{Setup}
\label{sec:exp:setup}

\par\medskip
\noindent\textbf{Benchmarks.}\quad
We evaluate on seven public benchmarks that together cover the capabilities analyzed in Section~\ref{sec:astra-grounding-analysis}: referring under spatial constraints and reasoning (RefSpatial~\cite{zhou2025roborefer}, PointArena~\cite{pointarena}), multi-instance perception (PixMo-Count~\cite{deitke2024molmopixmoopenweights} and the counting subset of PointArena), spatial relations and free-space localization (RoboSpatial~\cite{robospatial}, Where2Place~\cite{yuan2024robopoint}), and functional and interaction grounding (RoboAfford~\cite{roboafford} and the affordance split of ShareRobot~\cite{robobrain}). Every answer is a point, a count, or a yes/no token. We report the official metric of each benchmark, the mean over all seven (Avg-7). We also evaluate on \emph{RobotGrounding}, our own benchmark of 2{,}591 questions in 14 types on held-out robot manipulation frames with human-annotated boxes, ranging from referring and counting to absent targets and cross-view correspondence between two cameras; its score is the macro average over question types.

\par\medskip
\noindent\textbf{Protocol.}\quad
All models are evaluated with the same harness and the same prompt per benchmark, at temperature $0.2$ with a 128-token output budget; PixMo-Count, whose answers can list many points, uses a 1{,}024-token budget. Closed models are queried through their APIs with the same questions and scorer.

\subsection{Main Results}
\label{sec:exp:main}

\begin{table*}[!htb]
\centering
\caption{Results on seven public grounding benchmarks and on our RobotGrounding benchmark under one protocol (Section~\ref{sec:exp:setup}). Avg-7 averages the seven public benchmarks; RobotGrounding is not included in the average. Best in \textbf{bold}, second best \underline{underlined}.}
\label{tab:main}
\small
\setlength{\tabcolsep}{4pt}
\resizebox{\textwidth}{!}{%
\begin{tabular}{lccccccccc}
\toprule
Model & Where2Place & PixMo-Count & RoboSpatial & RefSpatial & PointArena & RoboAfford & ShareRobot & Avg-7 & RobotGrounding \\
\midrule
\multicolumn{10}{l}{\emph{Closed models}} \\
Claude Opus 5.5~\cite{anthropic2026opus55}
& 66.44 & 83.74 & 73.30 & 71.10 & 61.20 & 75.10 & 40.20 & 67.30 & 90.89 \\
Gemini 4 Argon~\cite{googledeepmind2026gemini4argon}
& 71.20 & 87.71 & \textbf{80.94} & 51.56 & 66.40 & 71.48 & 29.53 & 65.55 & 78.51 \\
GPT-5.5~\cite{openai2026gpt55}
& 45.25 & 82.61 & 59.19 & 60.29 & 68.02 & 68.04 & 38.57 & 60.28 & 83.96 \\
GPT-5.6-sol~\cite{openai2026gpt56sol}
& 46.23 & 83.55 & 64.34 & 54.51 & 69.76 & 71.45 & 36.66 & 60.93 & 85.18 \\
GPT-6 Astra~\cite{openai2026gpt6astra}
& 71.10 & \underline{88.10} & 69.55 & \textbf{80.14} & \textbf{75.36} & \underline{84.94} & 46.60 & \underline{73.68} & 91.10 \\
\midrule
\multicolumn{10}{l}{\emph{General-purpose open models}} \\
Qwen3-VL-8B~\cite{bai2025qwen3vltechnicalreport}
& 62.60 & 86.40 & 62.66 & 54.15 & 61.81 & 74.83 & 35.80 & 62.61 & \underline{92.02} \\
Qwen3.5-9B~\cite{qwen3.5}
& 72.00 & \textbf{90.35} & 53.46 & 57.76 & 68.33 & 78.65 & 39.47 & 65.72 & 86.70 \\
\midrule
\multicolumn{10}{l}{\emph{Open embodied models}} \\
RynnBrain-8B~\cite{dang2026}
& 68.00 & 11.70 & 70.00 & 56.68 & 65.58 & 79.29 & 15.50 & 52.39 & 80.14 \\
RoboBrain 2.5-8B~\cite{robobrain2_5}
& 64.00 & 78.80 & 63.71 & 59.21 & 59.27 & 76.33 & \textbf{56.24} & 65.37 & 81.35 \\
ACE-Brain-0.5-8B~\cite{acebrain0}
& 70.70 & 85.40 & 58.30 & 53.43 & 68.02 & 75.37 & 34.38 & 63.66 & 66.91 \\
Embodied-R1.5-8B~\cite{yuan2026}
& \underline{77.00} & 81.70 & 68.96 & 57.04 & 71.38 & 83.43 & 37.21 & 68.10 & 75.24 \\
SenseNova-SI-1.1-8B~\cite{cai2026sensenovasi}
& 47.00 & 59.20 & 59.65 & 29.78 & 51.83 & 50.71 & 25.98 & 46.31 & 66.83 \\
PhysBrain 1.5-8B~\cite{physbrain15}
& 76.30 & 85.63 & 69.74 & 55.60 & 69.86 & 81.24 & 33.47 & 67.41 & 87.05 \\
\midrule
\textbf{ACE-Brain-0.6 (ours)}
& \textbf{81.44} & 86.96 & \underline{76.56} & \underline{71.84} & \underline{74.13} & \textbf{88.14} & \underline{48.82} & \textbf{75.41} & \textbf{93.42} \\
\bottomrule
\end{tabular}}
\end{table*}

\begin{figure*}[!htb]
\centering
\includegraphics[width=\textwidth]{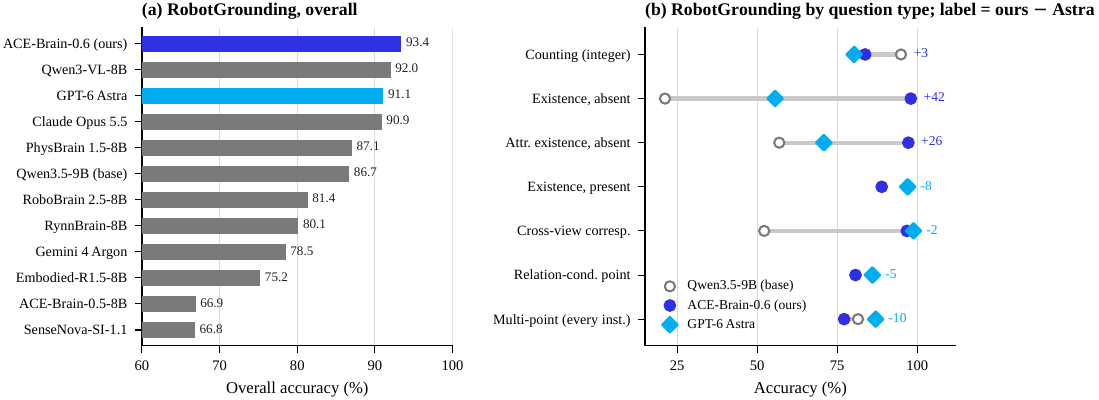}
\caption{ACE-Brain-0.6 and GPT-6 Astra on our robot-scene benchmark RobotGrounding. (a) Overall accuracy of all compared models. (b) Accuracy by question type for the base model, ours, and Astra; the gray bar connects the base model to ours, and the label is our score minus Astra's. Our model leads wherever the answer is a number, \emph{none}, or a correspondence across views; Astra leads when every instance must be enumerated or a relation resolved.}
\label{fig:capability}
\end{figure*}

\par\medskip
\noindent\textbf{Compared models.}\quad
We compare against Qwen3.5-9B~\cite{qwen3.5}; a general-purpose open VLM of the same size, Qwen3-VL-8B~\cite{bai2025qwen3vltechnicalreport}; open embodied models of the same scale, RynnBrain-8B~\cite{dang2026}, RoboBrain~2.5-8B~\cite{robobrain2_5}, ACE-Brain-0.5-8B~\cite{acebrain0}, Embodied-R1.5-8B~\cite{yuan2026}, SenseNova-SI-1.1-8B~\cite{cai2026sensenovasi}, and PhysBrain~1.5-8B~\cite{physbrain15}; and five closed models, Claude Opus~5.5~\cite{anthropic2026opus55}, Gemini~4 Argon~\cite{googledeepmind2026gemini4argon}, GPT-5.5~\cite{openai2026gpt55}, GPT-5.6-sol~\cite{openai2026gpt56sol} and GPT-6 Astra~\cite{openai2026gpt6astra}, the reference analyzed in Section~\ref{sec:astra-grounding-analysis}.

Table~\ref{tab:main} gives the full comparison, and Figure~\ref{fig:capability} details the results on RobotGrounding. Three results stand out: ACE-Brain-0.6 is the strongest open model of its size, it matches GPT-6 Astra on the seven-benchmark average while still trailing clearly on RefSpatial, and it leads every compared model in robot scenes.

Fine-tuning improves the base model on six of the seven benchmarks and raises Avg-7 by almost ten points. The largest gains appear where grounding requires a decision rather than a glance, on the spatial relations of RoboSpatial and the constrained references of RefSpatial, whereas on counting, which the base model already handles well, fine-tuning brings no gain and costs a few points.

Among open embodied models of the same scale, ACE-Brain-0.6 is the best on every benchmark except ShareRobot and leads the next model on Avg-7 by more than seven points. Against GPT-6 Astra the picture is mixed: our model is ahead on four benchmarks, slightly behind on PixMo-Count and PointArena, and well behind on RefSpatial (71.8 vs.\ 80.1), so the two averages end up close (Avg-7 75.4 vs.\ 73.7). The other four closed models trail ACE-Brain-0.6 on at least five of the seven benchmarks; only on RoboSpatial is Gemini~4 Argon ahead of every model. On RobotGrounding, ACE-Brain-0.6 is ahead of all compared models, including the five closed ones. Section~\ref{sec:exp:capability} relates these differences to the capabilities and limitations of Astra identified in Section~\ref{sec:astra-grounding-analysis}, using the labeled subsets of the public benchmarks (Table~\ref{tab:subsets}) and the question types of RobotGrounding.

\subsection{Where the Gains Come From}
\label{sec:exp:capability}

\begin{table}[!htb]
\centering
\caption{Scores of ACE-Brain-0.6 and GPT-6 Astra on the labeled subsets of four public benchmarks.}
\label{tab:subsets}
\small
\begin{tabular}{llcc}
\toprule
Benchmark & Subset & Ours & Astra \\
\midrule
RoboSpatial & Configuration (relation, yes/no)     & 83.74 & 86.2 \\
            & Compatibility (fits?, yes/no)        & 84.76 & 76.2 \\
            & Context (free point)                 & 62.27 & 53.4 \\
\midrule
RefSpatial  & Location                             & 73.00 & 77.0 \\
            & Placement                            & 75.00 & 84.0 \\
\midrule
PointArena  & Spatial                              & 85.64 & 88.2 \\
            & Reasoning                            & 75.13 & 56.3 \\
            & Counting (every instance)            & 63.27 & 60.0 \\
            & Affordable                           & 92.42 & 86.7 \\
\midrule
RoboAfford  & Object affordance (part)             & 91.94 & 84.7 \\
            & Object reference                     & 92.98 & 92.8 \\
            & Spatial affordance (free region)     & 77.92 & 70.8 \\
\bottomrule
\end{tabular}
\end{table}

\par\medskip
\noindent\textbf{Placement under spatial constraints.}\quad
Both models are accurate when a target is named by category, function, or attribute, and Astra is ahead on RefSpatial location, where the description adds a spatial constraint. The RefSpatial gap is larger in placement (Table~\ref{tab:subsets}), which tests the feasibility under spatial constraints discussed in Section~\ref{sec:astra-grounding-analysis}: the answer must satisfy a relation to one or more reference objects and lie in free space. A valid placement point also needs a supporting surface and enough clearance next to the reference. Our free-space targets mark unoccupied 2D regions and do not assert 3D feasibility (Section~\ref{sec:method:data}), so we expect the gap to require targets verified in 3D rather than more 2D free-space data. Consistent with this, adding several thousand placement questions with targets computed exactly from 2D object masks, a small share of the training data, left placement unchanged.

\par\medskip
\noindent\textbf{Parts and free space.}\quad
Our model leads Astra on all three RoboAfford subsets, on ShareRobot affordance, on RoboSpatial, and on free-space pointing in Where2Place. On RoboSpatial the lead comes from the compatibility and context questions, while Astra remains slightly ahead on relation judgments (Table~\ref{tab:subsets}). Part-level targets such as ``the handle'' or ``where to hold'' belong to the fine-grained geometric localization that Section~\ref{sec:astra-grounding-analysis} identified as a limitation of Astra. Across our data releases, the largest ShareRobot gain coincides with the addition of part masks selected by the action verb of the query, which suggests that this supervision closed the gap to Astra.

\par\medskip
\noindent\textbf{Robot scenes.}\quad
ACE-Brain-0.6 leads every compared model on RobotGrounding Benchmark (Figure~\ref{fig:capability}(a)). Most of the gain over the base model comes from three abilities it largely lacks: recognizing that a named object is absent, recognizing that no object has the requested attribute, and matching an object marked in one camera to the same object in another (Figure~\ref{fig:capability}(b)). Our model is far ahead of Astra on the first two and on par with it on the third, while Astra stays ahead when every instance must be pointed to or the target is defined by a relation. The two models also err in opposite directions: Astra often reports objects that are absent, whereas ours more often misses objects that are present.

\subsection{Data Scaling}
\label{sec:exp:scaling}

\begin{figure*}[!htb]
\centering
\includegraphics[width=\textwidth]{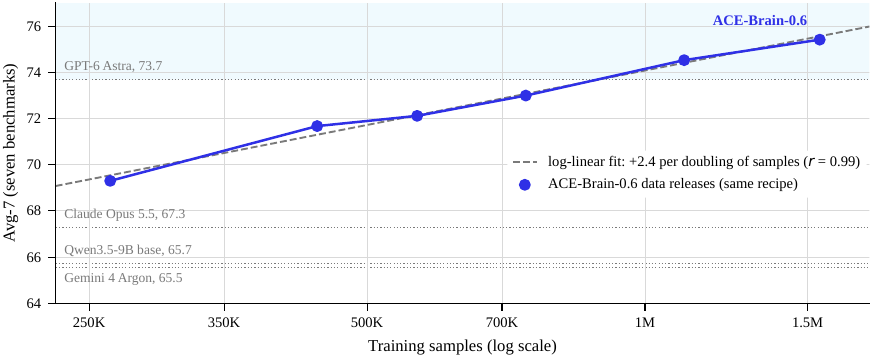}
\caption{Grounding performance against the number of training samples (log scale). Every point is a full training run from Qwen3.5-9B with the same recipe; only the data changes. Avg-7 over the data releases, with a log-linear fit; dotted lines mark the base model, GPT-6 Astra, Claude Opus~5.5, and Gemini~4 Argon.}
\label{fig:scaling}
\end{figure*}

The training collection of Section~\ref{sec:method:data} was assembled in a sequence of releases, from 0.26M to 1.55M samples, each adding new task families or image domains to the previous one. Every release was trained from the base model with the same recipe (Section~\ref{sec:method:training}), so the resulting checkpoints form a controlled series in which only the data differs. Figure~\ref{fig:scaling} plots them against the number of training samples.

\par\medskip
\noindent\textbf{Performance grows steadily with data and passes Astra’s average.}\quad
Over a roughly sixfold range of training data, from 0.26M to 1.55M samples, Avg-7 rises from 69.3 to 75.4 along a log-linear trend of about $+2.4$ points per doubling ($r = 0.99$); counting the base model, the full span is 65.7 to 75.4. The 1.55M release continues the trend, with no sign of saturation within the range we tested.

\par\medskip
\noindent\textbf{What scales is coverage, not repetition.}\quad
Each release added new kinds of supervision rather than more of the same, so in the main series the number of samples and the coverage of supervision grow together and cannot be fully separated. A single control run separates them: it doubles only the sources of an early release, reaching the size of the next release without any new task or domain, and gains less than one point over its starting point, landing within one point of the release that added new sources instead. The same pattern holds inside the sequence. Adding more samples whose answers list many points did not improve PointArena questions with many instances. For a fixed backbone and recipe, then, what moves grounding is the number of distinct kinds of supervision and how their answers are represented, not the number of examples.

\section{Conclusions}
\label{sec:conclusion}

We asked how far training data alone can carry an open 9B vision-language model toward the grounding capabilities of GPT-6 Astra. By collecting open robot, egocentric, object-centric, and spatial data, annotating the robot recordings automatically, and converting every source into question--answer supervision that can be verified against its evidence, we trained ACE-Brain-0.6 from Qwen3.5-9B with a standard single-epoch recipe. The model is ahead of Astra on four of seven public grounding benchmarks, on par on two, and above it on their average, and reaches 93.4 on our robot-scene benchmark against Astra's 91.1. A controlled series of training runs shows that grounding performance rises log-linearly with the number of training samples and that the rise is driven by the coverage of supervision.

Two limitations remain, and they mark the direction of the next release. ACE-Brain-0.6 still trails Astra when every instance of a set must be enumerated, and on placement under spatial constraints, where our two-dimensional free-space supervision does not capture three-dimensional feasibility. Both are properties of the data, and the scaling results suggest that the remedy is new supervision, complete multi-instance targets and feasibility-aware placement data, rather than more of what the collection already contains. These are the next additions to the pipeline.

\clearpage
\bibliographystyle{unsrtnat}
\bibliography{references}

@misc{openai2026gpt6astra,
  author = {{OpenAI}},
  title  = {{GPT-6 Astra}: A New Generation of Intelligence},
  year   = {2026},
  month  = sep,
  url    = {https://openai.com/index/gpt-6-astra/}
}

@misc{qwen3.5,
  author = {{Qwen Team}},
  title  = {{Qwen3.5}: Towards Native Multimodal Agents},
  year   = {2026},
  month  = feb,
  url    = {https://qwen.ai/blog?id=qwen3.5}
}

@inproceedings{kamath2021mdetr,
  author    = {Kamath, Aishwarya and Singh, Mannat and LeCun, Yann and
               Synnaeve, Gabriel and Misra, Ishan and Carion, Nicolas},
  title     = {{MDETR}: Modulated Detection for End-to-End Multi-Modal
               Understanding},
  booktitle = {Proceedings of the IEEE/CVF International Conference on
               Computer Vision},
  pages     = {1780--1790},
  year      = {2021}
}

@inproceedings{shridhar2021cliport,
  author    = {Shridhar, Mohit and Manuelli, Lucas and Fox, Dieter},
  title     = {{CLIPort}: What and Where Pathways for Robotic Manipulation},
  booktitle = {Proceedings of the 5th Conference on Robot Learning},
  series    = {Proceedings of Machine Learning Research},
  volume    = {164},
  pages     = {894--906},
  publisher = {PMLR},
  year      = {2022},
  url       = {https://proceedings.mlr.press/v164/shridhar22a.html}
}

@inproceedings{liu2023gres,
  author    = {Liu, Chang and Ding, Henghui and Jiang, Xudong},
  title     = {{GRES}: Generalized Referring Expression Segmentation},
  booktitle = {Proceedings of the IEEE/CVF Conference on Computer Vision
               and Pattern Recognition},
  pages     = {23592--23601},
  year      = {2023}
}

@inproceedings{wang2021ocid,
  author    = {Wang, Ke-Jyun and Liu, Yun-Hsuan and Su, Hung-Ting and
               Wang, Jen-Wei and Wang, Yu-Siang and Hsu, Winston and
               Chen, Wen-Chin},
  title     = {{OCID}-Ref: A 3{D} Robotic Dataset With Embodied Language
               for Clutter Scene Grounding},
  booktitle = {Proceedings of the 2021 Conference of the North American
               Chapter of the Association for Computational Linguistics:
               Human Language Technologies},
  pages     = {5333--5338},
  publisher = {Association for Computational Linguistics},
  year      = {2021},
  doi       = {10.18653/v1/2021.naacl-main.419},
  url       = {https://aclanthology.org/2021.naacl-main.419/}
}

@inproceedings{carion2026sam,
  title     = {{SAM 3}: Segment Anything with Concepts},
  author    = {Carion, Nicolas and Gustafson, Laura and Hu, Yuan-Ting and Debnath, Shoubhik and Hu, Ronghang and Suris Coll-Vinent, Didac and Ryali, Chaitanya and Alwala, Kalyan Vasudev and Khedr, Haitham and Huang, Andrew and others},
  booktitle = {International Conference on Learning Representations (ICLR)},
  year      = {2026}
}

@article{patzold2025leveraging,
  title     = {Leveraging Vision-Language Models for Open-Vocabulary Instance Segmentation and Tracking},
  author    = {P{\"a}tzold, Bastian and Nogga, Jan and Behnke, Sven},
  journal   = {IEEE Robotics and Automation Letters},
  year      = {2025},
  publisher = {IEEE}
}

@article{bai2025qwen3vltechnicalreport,
  title   = {{Qwen3-VL} Technical Report},
  author  = {Shuai Bai and Yuxuan Cai and Ruizhe Chen and Keqin Chen and Xionghui Chen and Zesen Cheng and Lianghao Deng and Wei Ding and Chang Gao and Chunjiang Ge and Wenbin Ge and Zhifang Guo and Qidong Huang and Jie Huang and Fei Huang and Binyuan Hui and Shutong Jiang and Zhaohai Li and Mingsheng Li and Mei Li and Kaixin Li and Zicheng Lin and Junyang Lin and Xuejing Liu and Jiawei Liu and Chenglong Liu and Yang Liu and Dayiheng Liu and Shixuan Liu and Dunjie Lu and Ruilin Luo and Chenxu Lv and Rui Men and Lingchen Meng and Xuancheng Ren and Xingzhang Ren and Sibo Song and Yuchong Sun and Jun Tang and Jianhong Tu and Jianqiang Wan and Peng Wang and Pengfei Wang and Qiuyue Wang and Yuxuan Wang and Tianbao Xie and Yiheng Xu and Haiyang Xu and Jin Xu and Zhibo Yang and Mingkun Yang and Jianxin Yang and An Yang and Bowen Yu and Fei Zhang and Hang Zhang and Xi Zhang and Bo Zheng and Humen Zhong and Jingren Zhou and Fan Zhou and Jing Zhou and Yuanzhi Zhu and Ke Zhu},
  year    = {2025},
  journal = {arXiv preprint arXiv:2511.21631}
}

@inproceedings{robobrain,
  title     = {{RoboBrain}: A Unified Brain Model for Robotic Manipulation from Abstract to Concrete},
  author    = {Ji, Yuheng and Tan, Huajie and Shi, Jiayu and Hao, Xiaoshuai and Zhang, Yuan and Zhang, Hengyuan and Wang, Pengwei and Zhao, Mengdi and Mu, Yao and An, Pengju and others},
  booktitle = {Proceedings of the Computer Vision and Pattern Recognition Conference (CVPR)},
  pages     = {1724--1734},
  year      = {2025}
}

@article{robobrain2_5,
  title   = {{RoboBrain 2.5}: Depth in Sight, Time in Mind},
  author  = {Tan, Huajie and Zhou, Enshen and Li, Zhiyu and Xu, Yijie and Ji, Yuheng and Chen, Xiansheng and Chi, Cheng and Wang, Pengwei and Jia, Huizhu and Ao, Yulong and others},
  journal = {arXiv preprint arXiv:2601.14352},
  year    = {2026}
}

@inproceedings{robospatial,
  title     = {{RoboSpatial}: Teaching Spatial Understanding to {2D} and {3D} Vision-Language Models for Robotics},
  author    = {Song, Chan Hee and Blukis, Valts and Tremblay, Jonathan and Tyree, Stephen and Su, Yu and Birchfield, Stan},
  booktitle = {Proceedings of the Computer Vision and Pattern Recognition Conference (CVPR)},
  pages     = {15768--15780},
  year      = {2025}
}

@inproceedings{roboafford,
  title     = {{RoboAfford}: A Dataset and Benchmark for Enhancing Object and Spatial Affordance Learning in Robot Manipulation},
  author    = {Tang, Yingbo and Zhang, Lingfeng and Zhang, Shuyi and Zhao, Yinuo and Hao, Xiaoshuai},
  booktitle = {Proceedings of the 33rd ACM International Conference on Multimedia},
  pages     = {12706--12713},
  year      = {2025}
}

@inproceedings{driess2023palme,
  title     = {{PaLM-E}: An Embodied Multimodal Language Model},
  author    = {Driess, Danny and others},
  booktitle = {International Conference on Machine Learning (ICML)},
  year      = {2023}
}

@inproceedings{yuan2024robopoint,
  title     = {{RoboPoint}: A Vision-Language Model for Spatial Affordance Prediction in Robotics},
  author    = {Yuan, Wentao and Duan, Jiafei and Blukis, Valts and Pumacay, Wilbert and Krishna, Ranjay and Murali, Adithyavairavan and Mousavian, Arsalan and Fox, Dieter},
  booktitle = {Proceedings of the 8th Conference on Robot Learning (CoRL)},
  series    = {Proceedings of Machine Learning Research},
  volume    = {270},
  pages     = {4005--4020},
  publisher = {PMLR},
  year      = {2025}
}

@article{zhou2025roborefer,
  title   = {{RoboRefer}: Towards Spatial Referring with Reasoning in Vision-Language Models for Robotics},
  author  = {Zhou, Enshen and An, Jingkun and Chi, Cheng and Han, Yi and Rong, Shanyu and Zhang, Chi and Wang, Pengwei and Wang, Zhongyuan and Huang, Tiejun and Sheng, Lu and Zhang, Shanghang},
  journal = {Advances in Neural Information Processing Systems},
  volume  = {38},
  year    = {2025}
}

@misc{deitke2024molmopixmoopenweights,
  title         = {{Molmo} and {PixMo}: Open Weights and Open Data for State-of-the-Art Vision-Language Models},
  author        = {Matt Deitke and Christopher Clark and Sangho Lee and Rohun Tripathi and Yue Yang and Jae Sung Park and Mohammadreza Salehi and Niklas Muennighoff and Kyle Lo and Luca Soldaini and Jiasen Lu and Taira Anderson and Erin Bransom and Kiana Ehsani and Huong Ngo and YenSung Chen and Ajay Patel and Mark Yatskar and Chris Callison-Burch and Andrew Head and Rose Hendrix and Favyen Bastani and Eli VanderBilt and Nathan Lambert and Yvonne Chou and Arnavi Chheda and Jenna Sparks and Sam Skjonsberg and Michael Schmitz and Aaron Sarnat and Byron Bischoff and Pete Walsh and Chris Newell and Piper Wolters and Tanmay Gupta and Kuo-Hao Zeng and Jon Borchardt and Dirk Groeneveld and Crystal Nam and Sophie Lebrecht and Caitlin Wittlif and Carissa Schoenick and Oscar Michel and Ranjay Krishna and Luca Weihs and Noah A. Smith and Hannaneh Hajishirzi and Ross Girshick and Ali Farhadi and Aniruddha Kembhavi},
  year          = {2024},
  eprint        = {2409.17146},
  archiveprefix = {arXiv},
  primaryclass  = {cs.CV},
  url           = {https://arxiv.org/abs/2409.17146}
}

@article{acebrain0,
  title   = {{ACE-Brain-0}: Spatial Intelligence as a Shared Scaffold for Universal Embodiments},
  author  = {Gong, Ziyang and Luo, Zehang and Tang, Anke and Liu, Zhe and Fu, Shi and Hou, Zhi and Yang, Ganlin and Wang, Weiyun and Wang, Xiaofeng and Liu, Jianbo and Luo, Gen and Kang, Haolan and Luo, Shuang and Zhou, Yue and Luo, Yong and Shen, Li and Jia, Xiaosong and Mu, Yao and Yang, Xue and Liu, Chunxiao and Yan, Junchi and Zhao, Hengshuang and Tao, Dacheng and Wang, Xiaogang},
  journal = {arXiv preprint arXiv:2603.03198},
  year    = {2026}
}

@misc{pointarena,
  title         = {{PointArena}: Probing Multimodal Grounding Through Language-Guided Pointing},
  author        = {Long Cheng and Jiafei Duan and Yi Ru Wang and Haoquan Fang and Boyang Li and Yushan Huang and Elvis Wang and Ainaz Eftekhar and Jason Lee and Wentao Yuan and Rose Hendrix and Noah A. Smith and Fei Xia and Dieter Fox and Ranjay Krishna},
  year          = {2025},
  eprint        = {2505.09990},
  archiveprefix = {arXiv},
  primaryclass  = {cs.CV},
  url           = {https://arxiv.org/abs/2505.09990}
}

@misc{li2022,
  title = {{Grounded Language-Image Pre-training}},
  author = {Liunian Harold Li and Pengchuan Zhang and Haotian Zhang and Jianwei Yang and Chunyuan Li and Yiwu Zhong and Lijuan Wang and Lu Yuan and Lei Zhang and Jenq-Neng Hwang and Kai-Wei Chang and Jianfeng Gao},
  year = {2022},
  eprint = {2112.03857},
  archivePrefix = {arXiv},
  primaryClass = {cs.CV},
  url = {https://arxiv.org/abs/2112.03857}
}

@misc{liu2024,
  title = {{Grounding DINO: Marrying DINO with Grounded Pre-Training for Open-Set Object Detection}},
  author = {Shilong Liu and Zhaoyang Zeng and Tianhe Ren and Feng Li and Hao Zhang and Jie Yang and Qing Jiang and Chunyuan Li and Jianwei Yang and Hang Su and Jun Zhu and Lei Zhang},
  year = {2024},
  eprint = {2303.05499},
  archivePrefix = {arXiv},
  primaryClass = {cs.CV},
  url = {https://arxiv.org/abs/2303.05499}
}

@misc{xiao2023,
  title = {{Florence-2: Advancing a Unified Representation for a Variety of Vision Tasks}},
  author = {Bin Xiao and Haiping Wu and Weijian Xu and Xiyang Dai and Houdong Hu and Yumao Lu and Michael Zeng and Ce Liu and Lu Yuan},
  year = {2023},
  eprint = {2311.06242},
  archivePrefix = {arXiv},
  primaryClass = {cs.CV},
  url = {https://arxiv.org/abs/2311.06242}
}

@inproceedings{chen2022pix2seq,
  title     = {{Pix2Seq}: A Language Modeling Framework for Object Detection},
  author    = {Chen, Ting and Saxena, Saurabh and Li, Lala
               and Fleet, David J. and Hinton, Geoffrey},
  booktitle = {International Conference on Learning Representations},
  year      = {2022},
  url       = {https://arxiv.org/abs/2109.10852}
}

@article{geminirobotics2025,
  title   = {{Gemini Robotics}: Bringing {AI} into the Physical World},
  author  = {{Gemini Robotics Team} and others},
  journal = {arXiv preprint arXiv:2503.20020},
  year    = {2025},
  url     = {https://arxiv.org/abs/2503.20020}
}

@misc{jiang2025,
  title = {{Detect Anything via Next Point Prediction}},
  author = {Qing Jiang and Junan Huo and Xingyu Chen and Yuda Xiong and Zhaoyang Zeng and Yihao Chen and Tianhe Ren and Junzhi Yu and Lei Zhang},
  year = {2025},
  eprint = {2510.12798},
  archivePrefix = {arXiv},
  primaryClass = {cs.CV},
  url = {https://arxiv.org/abs/2510.12798}
}

@misc{dang2026,
  title = {{RynnBrain: Open Embodied Foundation Models}},
  author = {Ronghao Dang and Jiayan Guo and Bohan Hou and Sicong Leng and Kehan Li and Xin Li and Jiangpin Liu and Yunxuan Mao and Zhikai Wang and Yuqian Yuan and Minghao Zhu and Xiao Lin and Yang Bai and Qian Jiang and Yaxi Zhao and Minghua Zeng and Junlong Gao and Yuming Jiang and Jun Cen and Siteng Huang and Liuyi Wang and Wenqiao Zhang and Chengju Liu and Jianfei Yang and Shijian Lu and Deli Zhao},
  year = {2026},
  eprint = {2602.14979},
  archivePrefix = {arXiv},
  primaryClass = {cs.RO},
  url = {https://arxiv.org/abs/2602.14979}
}

@misc{yuan2026,
  title = {{Embodied-R1.5: Evolving Physical Intelligence via Embodied Foundation Models}},
  author = {Yifu Yuan and Yaoting Huang and Xianze Yao and Yutong Li and Shuoheng Zhang and Linqi Han and Pengyi Li and Jiangeng Sun and Wenting Jia and Zhao Zhang and Yuhao Liu and Ruihao Liao and Yucheng Hu and Qiyu Wu and Yuxiao Li and Zibin Dong and Fei Ni and Yan Zheng and Shuyang Gu and Yi Ma and Hongyao Tang and Han Hu and Jianye Hao},
  year = {2026},
  eprint = {2606.11324},
  archivePrefix = {arXiv},
  primaryClass = {cs.RO},
  url = {https://arxiv.org/abs/2606.11324}
}

@misc{peng2023,
  title = {{Kosmos-2: Grounding Multimodal Large Language Models to the World}},
  author = {Zhiliang Peng and Wenhui Wang and Li Dong and Yaru Hao and Shaohan Huang and Shuming Ma and Furu Wei},
  year = {2023},
  eprint = {2306.14824},
  archivePrefix = {arXiv},
  primaryClass = {cs.CL},
  url = {https://arxiv.org/abs/2306.14824}
}

@misc{chen2024,
  title = {{SpatialVLM: Endowing Vision-Language Models with Spatial Reasoning Capabilities}},
  author = {Boyuan Chen and Zhuo Xu and Sean Kirmani and Brian Ichter and Danny Driess and Pete Florence and Dorsa Sadigh and Leonidas Guibas and Fei Xia},
  year = {2024},
  eprint = {2401.12168},
  archivePrefix = {arXiv},
  primaryClass = {cs.CV},
  url = {https://arxiv.org/abs/2401.12168}
}

@inproceedings{cai2026sensenovasi,
  title     = {Scaling Spatial Intelligence with Multimodal Foundation Models},
  author    = {Cai, Zhongang and Wang, Ruisi and Gu, Chenyang and Pu, Fanyi and Xu, Junxiang and Wang, Yubo and Yin, Wanqi and Yang, Zhitao and Wei, Chen and Zhou, Tongxi and Sun, Qingping and Pang, Hui En and Li, Jiaqi and Qian, Oscar and Lin, Zhiqian and Shi, Xuanke and Deng, Kewang and Han, Xiaoyang and Chen, Zukai and Fan, Xiangyu and Deng, Hanming and Lu, Lewei and Pan, Liang and Li, Bo and Liu, Ziwei and Wang, Quan and Lin, Dahua and Yang, Lei},
  booktitle = {Proceedings of the IEEE/CVF Conference on Computer Vision and Pattern Recognition (CVPR)},
  pages     = {7879--7890},
  year      = {2026}
}

@article{physbrain15,
  title   = {{PhysBrain 1.5}: From Vision-Language Models to Physical Foundation Models},
  author  = {{DeepCybo Team} and Bin, Yu and Cao, Haipeng and Chang, Zheng and Chen, Kai and Chen, Youning and Deng, Kailin and Du, Yichao and Fu, Xiaotong and Ge, Haoyang and others},
  journal = {arXiv preprint arXiv:2609.14973},
  year    = {2026}
}

@misc{anthropic2026opus55,
  title        = {System Card: {Claude Opus 5.5}},
  author       = {{Anthropic}},
  month        = {September},
  year         = {2026},
  howpublished = {\url{https://www.anthropic.com/model-card}}
}

@misc{googledeepmind2026gemini4argon,
  author       = {{Google DeepMind}},
  title        = {{Gemini 4 Argon}: Model Card},
  year         = {2026},
  month        = oct,
  url          = {https://deepmind.google/models/model-cards/gemini-4-argon/},
  note         = {Published October 13, 2026. Accessed October 17, 2026}
}

@misc{openai2026gpt55,
  author = {{OpenAI}},
  title  = {{GPT-5.5 Model}},
  year   = {2026},
  url    = {https://developers.openai.com/api/docs/models/gpt-5.5},
  note   = {OpenAI API documentation. Accessed: 2026-10-18}
}

@misc{openai2026gpt56sol,
  author = {{OpenAI}},
  title  = {{GPT-5.6 Sol Model}},
  year   = {2026},
  url    = {https://developers.openai.com/api/docs/models/gpt-5.6-sol},
  note   = {OpenAI API documentation. Accessed: 2026-10-18}
}

\clearpage
\appendix
\section{Appendix}
\subsection{Contributions and Author List}
\label{app:contributors}

\noindent\textbf{Contributors:}
Yinhe Wang\textsuperscript{1,*},
Tao Tao\textsuperscript{1,2,*},
Zexin He\textsuperscript{1,3},
Shuang Luo\textsuperscript{1,3},
Wei Mao\textsuperscript{1},
Yao Mu\textsuperscript{5},
Feng Pan\textsuperscript{1},
Hanming Deng\textsuperscript{4},
Lewei Lu\textsuperscript{4}.

\medskip
\noindent\textbf{Advisors:}
Zhi Hou\textsuperscript{1},
Xingang Pan\textsuperscript{3},
Dahua Lin\textsuperscript{4},
Dacheng Tao\textsuperscript{3},
Xiaogang Wang\textsuperscript{1}.

\medskip
\noindent\textbf{Affiliations:}\\
\textsuperscript{1}ACE Robotics\\
\textsuperscript{2}Sichuan University\\
\textsuperscript{3}Nanyang Technological University\\
\textsuperscript{4}SenseTime Research\\
\textsuperscript{5}Shanghai Jiao Tong University

\medskip
{\footnotesize
\noindent
\textsuperscript{*} Equal contribution.\par
}

\subsection{The RobotGrounding Benchmark}
\label{app:robotgrounding}

\begin{figure*}[!htb]
  \centering
  \includegraphics[width=\textwidth]{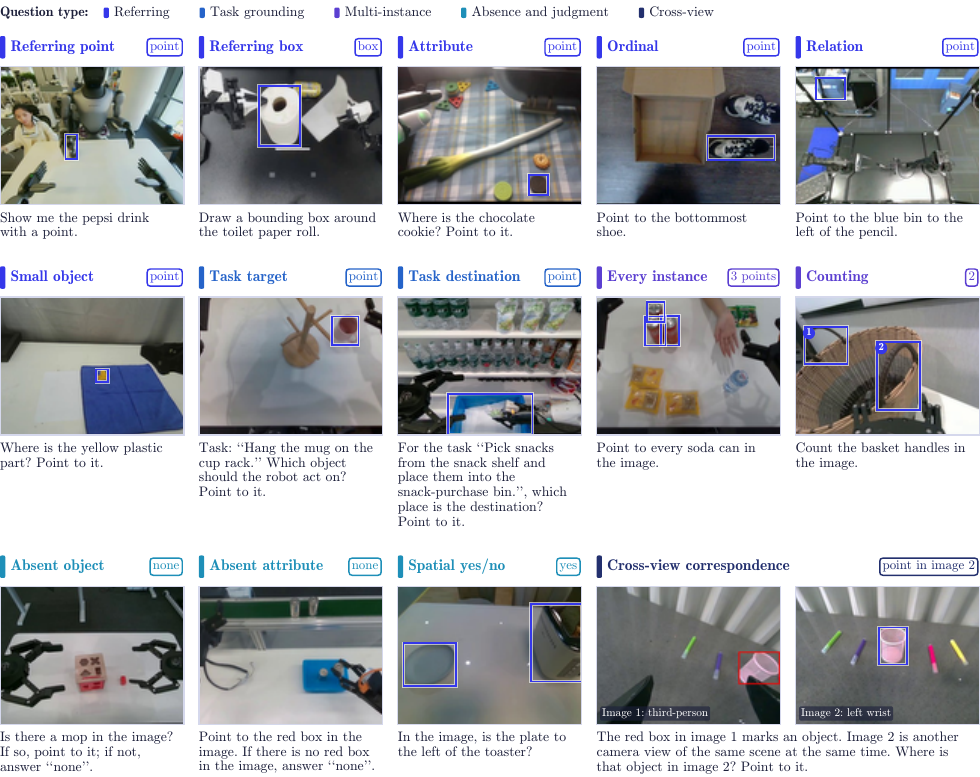}
  \caption{\textbf{RobotGrounding examples.} One question of each type, grouped by capability. Each panel shows the
  benchmark frame (letterbox bars removed; third-person camera unless marked), the question as posed to the model
  without the shared output-format instruction, and the expected answer form (top right). The blue box
  marks the task-relevant object, i.e., the reference answer (for the yes/no judgment, the two objects compared);
  absent-target questions have no box. The red box in the cross-view example is part of the input image.
  A point answer is correct when it falls inside or on the boundary of the reference box, a box answer at IoU $\geq 0.5$, and an
  every-instance answer requires exactly one point per instance, inside or on the boundary of its corresponding reference box.}
  \label{fig:rg_examples}
\end{figure*}

\begin{figure*}[!htb]
  \centering
  \includegraphics[width=\textwidth]{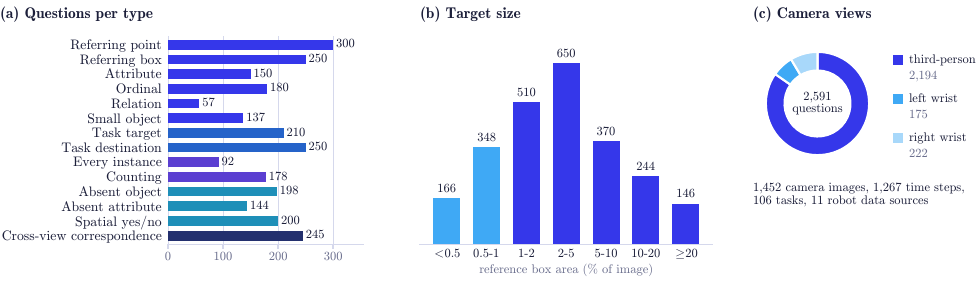}
  \caption{\textbf{RobotGrounding statistics.} (a) Questions per type, coloured by capability group as in
  Figure~\ref{fig:rg_examples}. (b) Size of the reference target, measured as the
  area of its box relative to the image; 21\% of the targets cover less than 1\% of the image. (c) Camera views;
  wrist-camera questions make up 15\% of the benchmark.}
  \label{fig:rg_stats}
\end{figure*}

RobotGrounding is our own evaluation set for grounding in the scenes a robot policy observes. It contains 2{,}591 questions over held-out frames from robot manipulation recordings, with object boxes annotated by hand and checked question by question; the episodes these frames come from are excluded from training, together with every other evaluation benchmark (Section~\ref{sec:method:data}). The questions follow the task families of Section~\ref{sec:method:model}: referring to an object by name, attribute, ordinal position, relation, or task role, answered with a box or a point; pointing to small objects that occupy well under one percent of the image; pointing to every instance of a described set; counting with an integer answer; existence questions whose correct answer is \emph{none} when the named object, or an object with the named attribute, is absent, paired with positive questions on objects that are present; spatial yes/no judgments; and cross-view correspondence, in which an object is marked by a box in one camera and must be located in a simultaneous view from another camera, without being named. Negative and positive questions are balanced within each type so that always answering \emph{none} does not score. Figure~\ref{fig:rg_examples} shows one example of each question type and Figure~\ref{fig:rg_stats} the composition of the benchmark.

A point answer is correct when it contains at least one point and every point lies inside or on the boundary of the target's annotated bounding box; an every-instance answer must contain exactly as many points as there are instances, each inside or on the boundary of a different annotated instance box; a box is correct at an intersection-over-union of at least 0.5; a count or a categorical answer must match exactly; and an absent-target question is answered correctly by any reply that contains no point. Because the reference regions are boxes rather than masks, the point criterion is lenient for thin or hollow objects. The overall score is the unweighted mean over 14 scoring groups, so that the many referring questions do not dominate the fewer, harder ones: referring by name with a point; referring with a box at IoU $\geq 0.5$ and, as a separate group, at IoU $\geq 0.75$; referring by attribute; by ordinal position or relation, merged into one group; by task role; placement locations; spatial yes/no judgments; pointing to every instance; existence; attribute existence; integer counting; small objects; and cross-view correspondence.

\clearpage

\end{document}